\documentclass[lettersize,journal]{IEEEtran}
\usepackage{amsmath,amsfonts}
\usepackage{algorithmic}
\usepackage{array}
\usepackage[caption=false,font=normalsize,labelfont=sf,textfont=sf]{subfig}
\usepackage{textcomp}
\usepackage{stfloats}
\usepackage{url}
\usepackage{verbatim}
\usepackage{graphicx}
\usepackage{multirow} 
\usepackage{rotating}
\usepackage{booktabs}
\usepackage{lipsum}
\usepackage{orcidlink}
\usepackage[table]{xcolor}

\def\BibTeX{{\rm B\kern-.05em{\sc i\kern-.025em b}\kern-.08em
    T\kern-.1667em\lower.7ex\hbox{E}\kern-.125emX}}
\usepackage{balance}
\begin{document}
\title{Distributed Acoustic Sensing for Urban Traffic Monitoring: Spatio-Temporal Attention in Recurrent Neural Networks}
\author{
    \IEEEauthorblockN{
    Izhan Fakhruzi\orcidlink{0000-0002-6968-4477},
    Manuel Titos\orcidlink{0000-0002-8279-2341},
    Carmen Benítez\orcidlink{0000-0002-5407-8335},
    Luz García\orcidlink{0000-0001-5904-5412}
    }
    \thanks{This work was supported in part by the grant no. PID2023-1523010B-IOO (MuSTARD), funded by the Agencia Estatal de Investigación in Spain, call no. MICIU/AEI/10.13039/501100011033, and in part by Junta de Andalucía/Consejería de Universidad, Investigación e Innovación/Proyecto P21\_00051 and by the generous collaboration of TM Digital™.\\
    \indent Izhan Fakhruzi, Manuel Titos, Carmen Benítez, and Luz García are with the Department of Signal Theory, Telematics and Communications, and the Center for Information and Communication Technologies Research (CITIC), University of Granada, 18014 Granada, Spain (emails: izhan@ugr.es/luzgm@ugr.es)\\
    \indent This work has been submitted to the IEEE for possible publication. Copyright may be transferred without notice, after which this version may no longer be accessible.}
}

\markboth{Submitted to IEEE Transactions on Intelligent Transportation Systems (T-ITS), March~2026.}%
{How to Use the IEEEtran \LaTeX \ Templates}

\maketitle

\begin{abstract}
Effective urban traffic monitoring is essential for improving mobility, enhancing safety, and supporting sustainable cities. Distributed Acoustic Sensing (DAS) enables large-scale traffic observation by transforming existing fiber-optic infrastructure into dense arrays of vibration sensors. However, modeling the high-resolution spatio-temporal structure of DAS data for reliable traffic event recognition remains challenging. This study presents a real-world DAS-based traffic monitoring experiment conducted in Granada, Spain, where vehicles cross a fiber deployed perpendicular to the roadway. Recurrent neural networks (RNNs) are employed to model intra- and inter-event temporal dependencies. Spatial and temporal attention mechanisms are systematically integrated within the RNN architecture to analyze their impact on recognition performance, parameter efficiency, and interpretability. Results show that an appropriate and complementary placement of attention modules improves the balance between accuracy and model complexity. Attention heatmaps provide physically meaningful interpretations of classification decisions by highlighting informative spatial locations and temporal segments. Furthermore, the proposed SA-bi-TA configuration demonstrates spatial transferability, successfully recognizing traffic events at sensing locations different from those used during training, with only moderate performance degradation. These findings support the development of scalable and interpretable DAS-based traffic monitoring systems capable of operating under heterogeneous urban sensing conditions.
\end{abstract}


\section{Introduction}
\label{sec:Int}
With rapidly increasing traffic volumes in urban areas, Intelligent Transportation Systems (ITS) have become essential for effective traffic management. By leveraging advanced information and communication technologies, ITS aim to reduce environmental impact, improve traffic flow and safety, and enhance emergency response, supporting the development of sustainable cities \cite{Elassy2024, Min2024}. Within this context, urban traffic monitoring enables continuous and automatic extraction of vehicle trajectories, speeds, and characteristics \cite{Liu2023, Chiang2024}. Various sensing technologies have been explored, including video-based \cite{Datondji2016}, pavement-embedded \cite{Goli2018}, and mobile phone-based systems \cite{Janecek2015}, demonstrating robust performance. However, these approaches are often limited by restricted spatial coverage, costly deployment requiring road closures, and privacy concerns \cite{Jain2019, Nuria2020}.\\
\indent Distributed Acoustic Sensing (DAS) \cite{Wu2025}, a fiber-optic–based sensing technology, leverages existing urban fiber infrastructures to enable dense, distributed traffic monitoring without installing dedicated sensors at multiple locations. DAS operates by launching optical pulses into the fiber and analyzing the Rayleigh backscattered light caused by microscopic inhomogeneities \cite{Fernandez-Ruiz2019}. This mechanism provides time-continuous measurements with high spatial resolution along the entire fiber span. DAS has been increasingly investigated for urban traffic monitoring due to its continuous and real-time sensing capability, wide coverage range, high sensitivity, immunity to electromagnetic interference, and minimal deployment requirements \cite{Hall2019, Li2021, Kandamali2022}.\\
\indent Despite DAS advantages, retrieving high-quality data from pre-existing fibers is affected by environmental factors \cite{Yang2021}, including ground properties, depth of fiber conduits, physical characteristics of fiber optic cables and the degree of fiber-conduit and conduit-ground coupling. These factors result in reduced generalizability of the methods when applied to heterogeneous conditions along fiber optic networks. To address these challenges and improve DAS sensing and recognition ability, researchers have explored various methodologies under signal processing, Machine Learning (ML), and Deep Learning (DL) frameworks \cite{Wu2025, Venketeswaran2022, Alashwal2023}. The process begins with advanced signal processing approaches, such as bandpass filtering \cite{Ye2023}, Spatial-Domain Bayesian filtering \cite{Liu2023}, improved wavelet-denoising or dual-threshold algorithms \cite{Liu2020}, essential for enhancing DAS signal quality and interpretability. After signal enhancement, some studies have focused on DAS event characterization relying on domain-informed handcrafted features such as Log-frequency Cepstral Coefficients \cite{Fei2018}, Hjorth parameters \cite{Luz2023}, or Hough Transforms combined with peak-to-peak amplitude descriptors \cite{Corera2023} among others. Prior to ML modeling, Exploratory Data Analysis \cite{Carlos2023, Fakhruzi2025} is often conducted to assess latent structures, class separability, relevant frequency bands, and feature importance. Events detection and classification is subsequently performed using algorithms such as Support Vector Machines (SVM) \cite{Li2021, Corera2023}, Random Forest \cite{Fakhruzi2025, Wang2019}, or Multi-layer Perceptron (MLP) \cite{Wu2017}. Despite their reported performance, these methods are not inherently designed for sequential modeling and therefore exhibit limited effectiveness on continuous records.\\
\indent Advancing toward effective automatic traffic event recognition using DAS requires models capable of capturing temporal dynamics. Early sequence modeling relied on traditional ML methods such as Hidden Markov Models (HMMs), which successfully extracted contextual information for pipeline monitoring \cite{Wu2019}. More recent works have extensively explored DL approaches, particularly Recurrent Neural Networks (RNNs), due to their ability to model temporal dependencies in DAS time series. Some studies employ handcrafted features processed by Long Short-Term Memory (LSTM), RNN variant, to identify seismic sequences \cite{Javi2025} or standardized anthropogenic and natural vibration events \cite{Qiao2025}. Others adopt end-to-end DL architectures in which Convolutional Neural Networks (CNNs) act as automatic feature extractors, while LSTM variants perform sequence recognition. This strategy has been applied to mechanical activity monitoring \cite{Wu2020} and railroad condition surveillance \cite{Rahman2024}.\\
\indent Recently, attention mechanisms---inspired by selective focus in the human visual system---have significantly improved DL architectures \cite{Bahdanau2015, Niu2021}. In the DAS domain, these strategies enable networks to emphasize salient spatial and temporal patterns within high-dimensional spatio-temporal DAS data. For instance, \cite{Chen2020} introduced a mechanism to capture critical segments of disturbance signals. Similarly, \cite{LiDong2024} integrated a multi-head attention module into a CNN-LSTM architecture to aggregate complementary representations, enhancing features robustness. Further advances have explored dual-attention mechanisms \cite{Tian2022, Li2024} integrating Channel Attention (CA) that acts as feature filter, and Spatial Attention (SA) highlighting relevant spatial points. Despite these advances in attention mechanisms and sequence modeling, most existing studies focus on isolated event sequences under highly controlled conditions. In real-world DAS applications, multiple and various events may occur sequentially or overlap within a signal segment. This study focuses on identifying continuous event sequences in real-time dynamic traffic settings.\\
\indent Within this context, this work tackles the challenging problem of continuous traffic event recognition in a real urban DAS-based monitoring deployment in Granada, Spain, where several fiber locations are continuously observed. A curated, high-quality real-world traffic dataset---manually labeled by domain experts---is constructed to investigate optimal strategies for modeling temporal dynamics. Spatio-temporal attention mechanisms, together with complementary approaches for extracting discriminative spatial and temporal features, are systematically evaluated. The main contributions of this study are as follows:
\begin{enumerate}
    \item To establish a real-world urban DAS-based traffic monitoring framework for continuous event recognition beyond isolated, controlled scenarios. We design denoising, preprocessing, and parameterization strategies that enable the construction of a high-value curated dataset and consolidate methodological expertise in the effective deployment of DAS for urban traffic analysis.
    \item To systematically evaluate model architectures and spatio-temporal attention strategies for modeling traffic event dynamics, explicitly accounting for dataset constraints such as limited size, high noise levels, and partial class separability.
    \item  To analyze the transferability of the proposed architectures across different sensing locations within a heterogeneous and operational urban fiber network.
\end{enumerate}

The remainder of this paper is organized as follows. Section \ref{sec:arc_exp} outlines the motivation for employing recurrent architectures and attention mechanisms. Section \ref{sec:exp_desc} details the experimental setup and reports the corresponding numerical results. For clarity, the subsequent sections analyze these results from complementary perspectives. Section \ref{sec:comparative-rnn} compares the evaluated RNN architectures and feature configurations. Section \ref{sec:ablation-study} presents the ablation study on the attention modules, while Section \ref{sec:att-weights} examines the resulting attention heatmaps. Section \ref{sec:spat-trans} investigates the spatial transferability of the proposed models across different sensing locations within the urban fiber deployment. Section \ref{sec:insights} provides a comprehensive discussion of the overall insights derived from the study. Finally, Section \ref{sec:conc} concludes the paper and outlines future research directions.

\section{Rationale for Model Architecture}
\label{sec:arc_exp}
Different alternatives exist and have been benchmarked \cite{Navid2024,Ismail2019, Wen2023} for time-series analysis, depending on the type of temporal structure under study (e.g., stationary vs. non-stationary, periodic vs. irregular), data availability or scarcity, and the target task (forecasting, classification, anomaly detection, representation learning, etc.). While convolutional (e.g., FCN/TCN) and Transformer-based models constitute strong alternatives for time-series event classification, in this work we focus on RNN architectures, exploring the potential benefits of incorporating attention mechanisms. Following the discussions in recent surveys, this model choice is motivated by: (i) the limited size of our labeled dataset; (ii) the need to capture temporally extended event structures while ensuring stable frame-wise predictions; (iii) the brief duration of vehicle-induced events and their temporal dynamics which can be effectively modeled without requiring long memory or large datasets, unlike architectures with large receptive fields or context windows such as TCNs or Transformers; and (iv) the discriminative power of the knowledge-based feature set used to parametrize the data (see Section \ref{sec:exp_desc}), which was already analyzed in a previous EDA work of our recordings \cite{Fakhruzi2025}.
\subsection{Recurrent Neural Networks for DAS time-series modeling}
\label{sec:rnn}
RNNs handle sequential tasks using recurrent connections \cite{Bengio1994}. These connections enable the model to retain information across time but suffer in longer sequences due to the vanishing and exploding gradients. Long Short-Term Memory (LSTM) was proposed to address these issues through gating mechanisms by regulating information flow \cite{Jurgen1997}. In addition, the activation maps of the LSTM hidden states provide strong interpretability, offering extra insight into how the LSTM models temporal information and decides its classification output \cite{Titos2022}. The effectiveness of LSTM has been successfully applied in numerous DAS applications including pipeline \cite{Yang2021}, seismicity\cite{Javi2025}, railway \cite{Rahman2024}, and intrusion \cite{LiDong2024} monitoring systems.\\
\indent Given the strengths of LSTM architectures, we employ an LSTM architecture to capture both long-term dependencies---modeling relationships across traffic events—--and short-term dependencies within neighboring window frames of the same event. Furthermore, its bidirectional variant (bi-LSTM) contains forward and backward layers, whose concatenated hidden states form the final sequence representation. This design preserves both past and future context, enabling a richer temporal encoding and more robust identification of continuous traffic event sequences.
\subsection{Attention mechanisms}
\label{sec:attention}
Attention mechanisms have emerged as a key component for improving the performance of deep neural networks in sequence modeling tasks. Inspired by the selective processing observed in the human visual system, attention enables models to focus on the most informative temporal segments of the input sequence, as well as on the most relevant spatial locations along the sensing fiber, rather than treating all time steps and spatial points uniformly.\\
\indent Originally proposed to enable dynamic alignment in sequence-to-sequence models, attention mechanisms have evolved into multiple variants characterized by their structural design (e.g., global vs. local, self- vs. cross-attention) and scoring functions (e.g., additive or multiplicative). This evolution culminated in the Transformer architecture \cite{Vaswani2017}, entirely based on self-attention and widely applied to time-series and spatio-temporal modeling.\\
\indent The attention mechanisms have been incorporated rapidly into diverse applications ranging from human action recognition \cite{Dai2020}, image caption generation \cite{Lu2017}, speech recognition \cite{Chan2016}, to urban traffic monitoring \cite{Reza2025}. In time series and sequential modeling, attention has emerged as an important component. For example, in DAS applications, attention mechanisms have been utilized for denoising and signal enhancement \cite{Lidenosing2023, Sui2023}, and event recognition \cite{Tian2022, Li2024}. The effectiveness of Transformers has also been explored in various DAS applications \cite{Cheng2024, Ding2025}.

    \begin{figure}[!ht]
        \center\includegraphics[width=6cm]{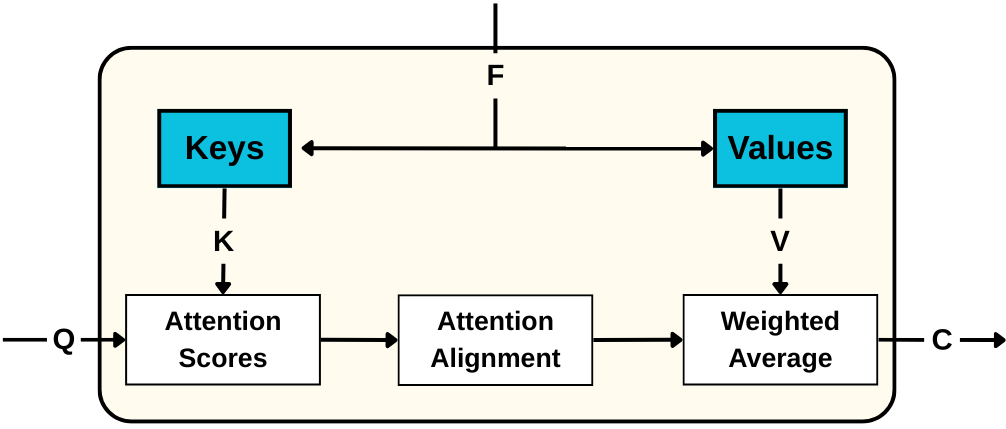}
        \caption{General attention module (adapted from \cite{Brauwers2023}).}
        \label{fig:attention}
    \end{figure}
    
\indent In general, an attention mechanism (see Fig. \ref{fig:attention}) involves two main steps: (i) computing attention weights (or alignment scores) that quantify the relevance between elements of the input, and (ii) forming a context vector as a weighted combination of input features, which is then used by the base model for subsequent processing. Most attention mechanisms follow a unified formulation based on three core components: Query ($Q$), Key ($K$), and Value ($V$). $Q$ represents the information to be retrieved, $K$ characterizes the content of the input elements, and $V$ contains the corresponding feature representations. $K$ and $V$ are typically derived from the input feature vectors $\mathbf{F}$. The resulting context vector $C$ aggregates the $V$ according to the computed attention weights and serves as an informative summary of the input. When $Q$, $K$ and $V$ are derived from the same input sequence, the mechanism is  referred to as self-attention, enabling the model to capture dependencies within the input sequence or spatial domain. In a self-attention mechanism, $Q$, $K$ and $V$ matrices are obtained through linear projections of the input embeddings. The similarity between queries and keys is computed via the dot product $QK^{T}$, producing attention scores that capture the representation of each time frame within the context of the entire sequence. These scores are scaled by the square root of the key dimensionality ($\sqrt{d_k}$) to stabilize gradients and prevent excessively large values (see Equation 1). This mechanism enables the modeling of contextual relationships both across spatial points (SPs) and across window frames in the DAS records, regardless of their relative positions within the sequence.
    \begin{align}
        \text{Attention Weights} &= \text{softmax}\left( \frac{QK^\top}{\sqrt{d_k}} \right) \label{eq:attn_weight}
    \end{align}
These attention weights are then multiplied by the $V$ (Equation \ref{eq:attn_output}), providing a dynamic representation of the input with higher weights assigned to more informative SPs or window frames.
    \begin{align}
        \text{Attention}(Q, K, V) &= \text{Attention Weights} \cdot V 
        \label{eq:attn_output}
    \end{align}
\indent Leveraging self-attention mechanisms to dynamically focus on different SPs and temporal windows within the DAS recordings, we evaluate the integration of spatio-temporal attention modules into the learning process of a bi-LSTM. Attention modules comprise spatial attention (SA) and temporal attention (TA), enabling to selectively emphasize the most informative spatial and temporal features for traffic event recognition. The experiments conducted in the remainder of this work investigate two key aspects of the potential spatio-temporal attention modules applicable to our data. First, we analyze whether the inclusion of attention mechanisms enhances the model’s ability to characterize traffic events. Second, we examine how different attention mechanisms (SA and TA) lead to distinct modeling effects together with the impact of their placement and ordering within the architecture, on overall robustness and performance.

\section{Experiment Description} 
\label{sec:exp_desc}
\subsection{Monitoring scenario}
\label{subsec:monitoringscen}
\begin{figure}
    \center\includegraphics[width=8.6cm]{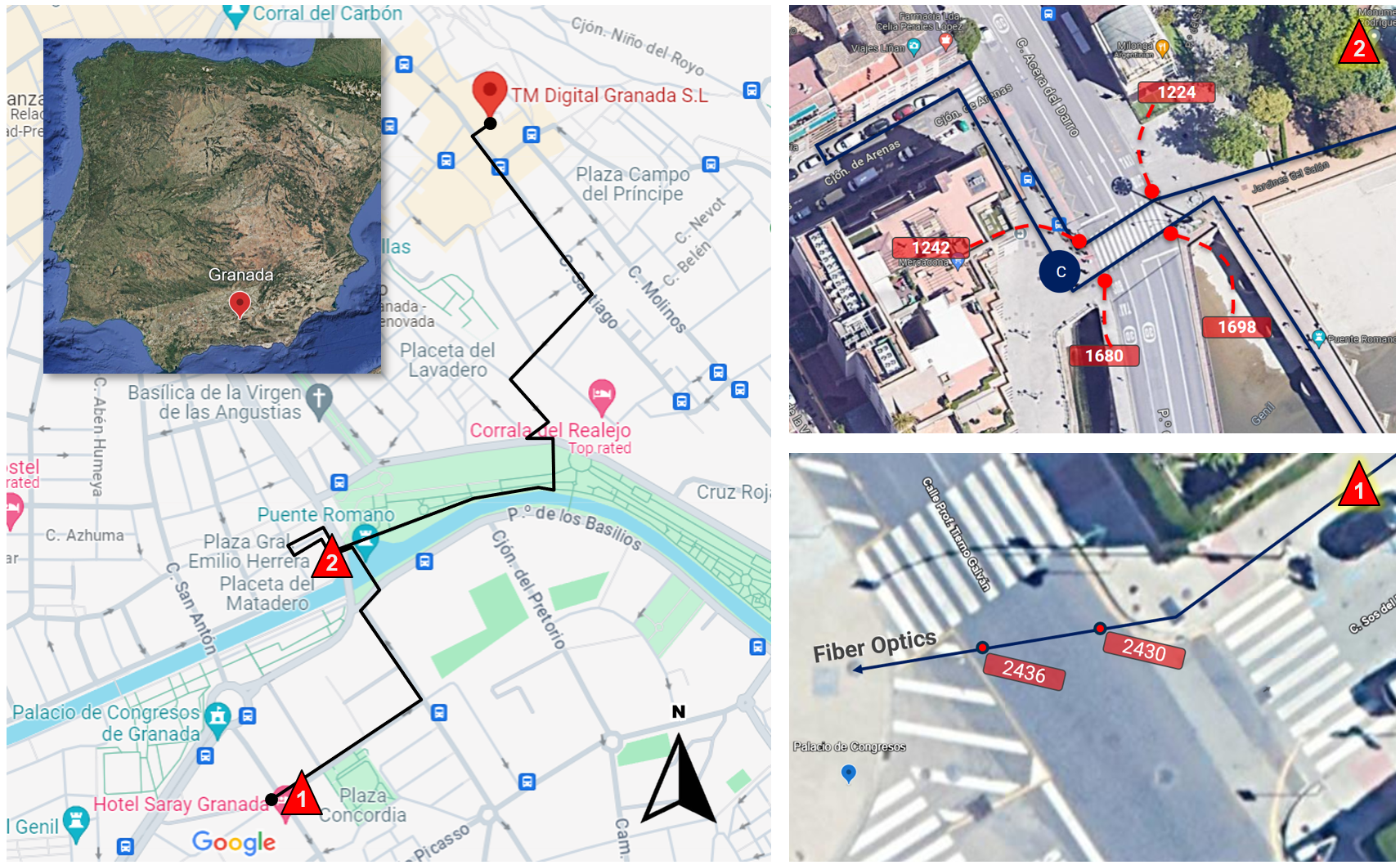}
    \caption{2.5 km of fiber, the red triangles represent the locations of the fiber where the registered data are collected and used in this study. (1) \textit{Palacio de Congresos}; (2) \textit{Acera del Darro} crossroads. (Courtesy of Google Maps)}
    \label{fig:das_map}
\end{figure}

The system used in this study is a High-Fidelity Distributed Acoustic Sensor (HDAS) from Aragón Photonics™, based on Chirped-Pulse Phase-Sensitive Optical Time-Domain Reflectometry (CP-$\phi$OTDR). It was deployed on existing telecommunication fiber infrastructure in Granada, Spain. The left side of Fig. \ref{fig:das_map} shows the 2.5-km fiber segment deployed across key locations in the city center. This fiber operates as a dense sensor array, capturing mechanical waves generated by events such as pedestrians, cars, trucks, and buses. Event footprints are recorded as strain-rate signals ($\Delta\varepsilon$) at a sampling rate of 250 Hz, forming a spatio-temporal matrix $\mathbf{X} \in \mathbb{R}^{s \times t}$, where $s$ denotes the number of spatial points (SPs) and $t$ the number of temporal samples. The spatial resolution is 6 m per SP, with a channel spacing of 3 m.\\
\indent For this study, we focus on two pedestrian crossings of particular relevance for real-time urban traffic monitoring: \textit{Palacio de Congresos} and \textit{Acera del Darro}. These intersections are marked by red triangles on the right side of Fig. 2, where the fiber runs perpendicular to the roadway, generating one-dimensional data recorded at selected high-quality SPs.

\subsubsection{Palacio de Congresos}
\begin{figure}
    \center\includegraphics[width=8.6cm]{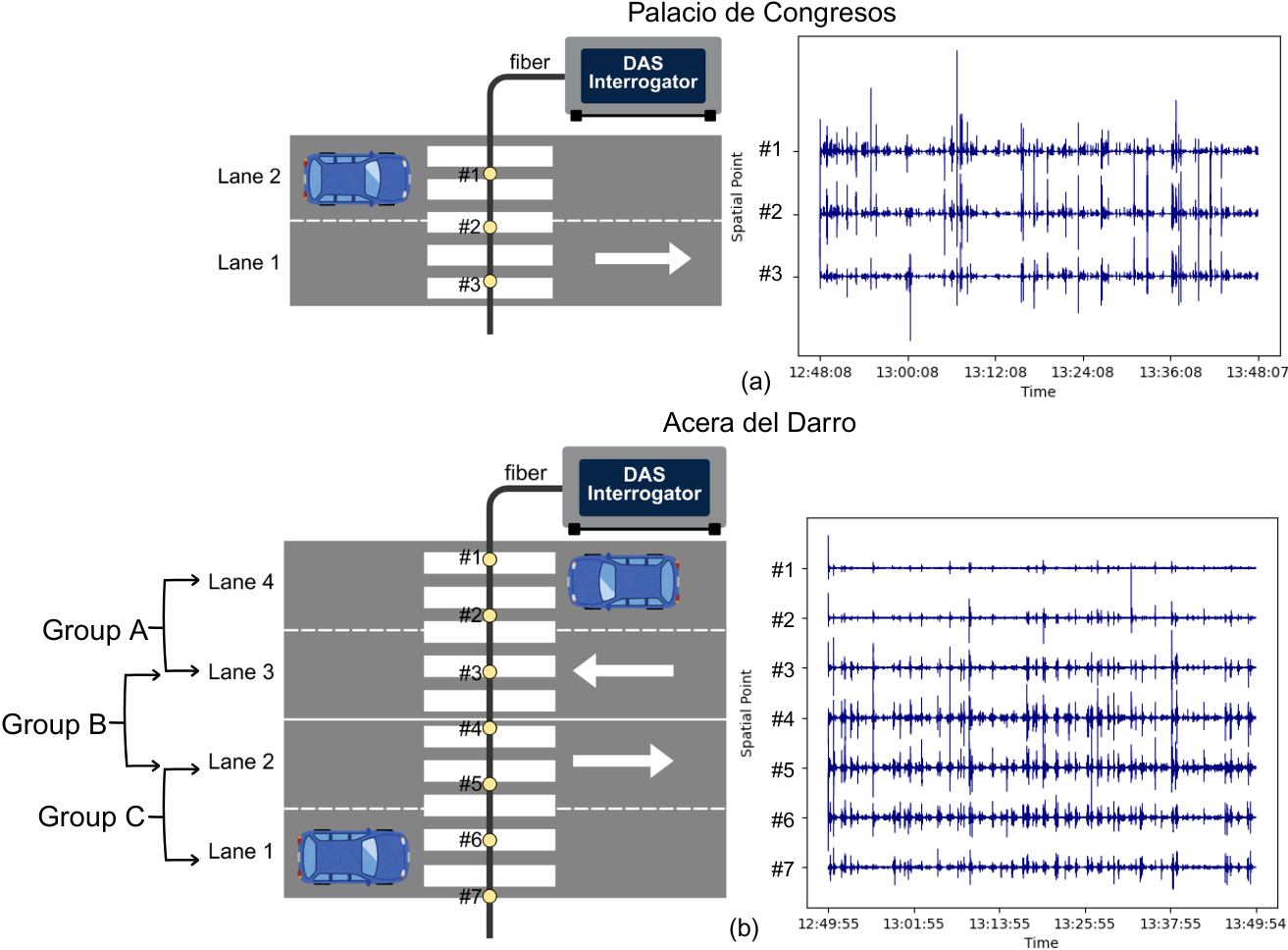}
    \caption{(Left) Schematic overview of fiber laid perpendicular to the lanes. Identified SPs on the fiber is denoted by \#1, \#2, ..., \#n. (Right) DAS registers of a 10-minute long signals with y-axis corresponds to the SPs. The fiber layout is depicted for: (a) \textit{Palacio de Congresos} with 2 lanes, 1 direction, and 3 identified SPs; (b) \textit{Acera del Darro} with 4 lanes, 2 directions, and 7 identified SPs, devided into 3 Groups of SPs: A(\#1--\#3), B(\#3--\#5), C(\#5--\#7).}
    \label{fig:layout-acera-palacio}
\end{figure}

This site comprises two unidirectional lanes. Following on-site inspection, three relevant SPs were identified to effectively capture traffic activity, as illustrated in Fig. \ref{fig:layout-acera-palacio}~(a).
\subsubsection{Acera del Darro}
In contrast, this intersection consists of four lanes arranged in two directions, leading to higher traffic volume and structural complexity. Seven effective SPs were selected to capture events across all lanes (see Fig. 3(b)). These SPs are grouped into three sets and used to evaluate the spatial transferability of the proposed model, as detailed in Section \ref{sec:spat-trans}.

\subsection{Dataset description}
\label{subsec:datadesc}
A total of 10 hours of observational data were collected at the \textit{Palacio de Congresos}, and 1 hour at \textit{Acera del Darro}. Based on observations from both sites, three canonical event categories were defined: \textit{Noise}, \textit{Car}, and \textit{Bus}. Video recordings and structured note-taking were conducted to register event types and their corresponding start and end times as vehicles passed the crossroads. The resulting annotations were subsequently synchronized with the DAS data to ensure accurate manual labeling.

\begin{table}[!ht]
    \caption{Summary of window counts (\#W) and event durations (Mean $\pm$ Std, in seconds (s))  generated from 10- and 1-hour observations at Palacio de Congresos and Acero del Darro Crossroads.\\}
    \centering
    \footnotesize
    \label{tab: urban_events}
    \scriptsize{
    \begin{tabular}{l p{0.8in} c c c}
        \hline
        \hline
        \textbf{Event}   & \textbf{Description}   & \textbf{\textit{Palacio.}(\#W)} &\textbf{\textit{Acera.}(\#W)} &\textbf{Duration (s)}\\
        \hline
        \textit{Noise}   & Pedestrian or no event    & 43,129   & 3,482 & 11.73$\pm$12.70\\
        \textit{Car}     & All types of cars     & 19,460   & 2,297 &5.72$\pm$3.05\\
        \textit{Bus}     & Large-size vehicles (all types of buses \& trucks)   & 9,411    & 1,421 &15.86$\pm$6.07\\
        \hline
        \multicolumn{2}{c}{Total}   & 72,000    & 7,200\\
        \hline
    \end{tabular}
    }
\end{table}

\indent Signal preprocessing and window-based analysis were performed to generate handcrafted features and aligned labels, following the procedure described in \cite{Fakhruzi2025}. The distribution of events and their mean durations at the monitoring sites is summarized in Table I. The workflow is outlined below:
\begin{table*}[ht]
\centering
\caption{Performance Comparison of the Models using feature Vectors with/without Temporal Derivatives ($X$ and $X$,+$\Delta$). Evaluation metrics include accuracy (Acc), F1-score (F1), number of trainable parameters (\#Param), relative improvement in accuracy over the baseline (RI-Acc), and relative increase in the number of trainable parameters with respect to the baseline (RPI), with $\pm$ indicating standard deviation across 5-fold cross-validation. The bi-LSTM model serves as the baseline for all comparisons highlighted in gray and \#Param is reported in millions (M).}
\label{tab:model_comparison}
\scriptsize{
\begin{tabular}{|l|ccccr|ccccr|}
\hline
\hline
\textbf{Model} & \multicolumn{5}{c|}{\textbf{$X$}} & \multicolumn{5}{c|}{\textbf{$X$,+$\Delta$}} \\ \cline{2-11} 
 & \textbf{Acc(\%)} & \textbf{F1(\%)} & \textbf{\#Param(M)} & \textbf{RI-Acc(\%)} & \textbf{RPI(\%)} 
 & \textbf{Acc(\%)} & \textbf{F1(\%)} & \textbf{\#Param(M)} & \textbf{RI-Acc(\%)} & \textbf{RPI(\%)} \\ \hline
LSTM    & 86.62$_{\pm0.02}$ &86.59$_{\pm0.02}$ &0.08 &-1.65 &-79.55
&88.17$_{\pm0.01}$ &88.18$_{\pm0.01}$ &0.92 &0.11 &124.35 \\
\hline
\rowcolor{gray!15}
bi-LSTM &88.08$_{\pm0.01}$ &88.10$_{\pm0.01}$ &0.41 &0.00 &0.00 
&88.38$_{\pm0.01}$ &88.41$_{\pm0.01}$ &0.69 &0.35 &68.09  \\
\hline
bi-TA   &88.47$_{\pm0.01}$ &88.49$_{\pm0.01}$ &0.25 &0.45 &-38.85
&88.13$_{\pm0.01}$ &88.11$_{\pm0.01}$ &3.36 &0.06 &715.96  \\
\hline
TA-bi   &87.97$_{\pm0.01}$ &88.06$_{\pm0.01}$ &0.32 &-0.12 &-23.21
&88.41$_{\pm0.00}$ &88.45$_{\pm0.01}$ &1.34 &0.37 &224.72  \\
\hline
bi-SA   &88.23$_{\pm0.01}$ &88.25$_{\pm0.01}$ &0.90 &0.18 &119.50
&88.02$_{\pm0.01}$ &88.02$_{\pm0.01}$ &0.92 &-0.06 &123.26  \\
\hline
SA-bi   &88.07$_{\pm0.01}$ &88.09$_{\pm0.01}$ &2.10 &-0.01 &408.64
&89.05$_{\pm0.01}$ &89.05$_{\pm0.01}$ &0.84 &1.10 &103.50  \\
\hline
SA-bi-TA &88.62$_{\pm0.01}$ &88.65$_{\pm0.01}$ &0.68 &0.62 &65.48
&89.05$_{\pm0.01}$ &89.03$_{\pm0.01}$ &1.13 &1.10 &174.15  \\
\hline
TA-bi-SA &87.24$_{\pm0.01}$ &87.28$_{\pm0.01}$ &0.72 &-0.95 &74.94
&88.20$_{\pm0.01}$ &88.22$_{\pm0.01}$ &2.70 &0.14 &555.32  \\
\hline
SA-TA-bi &87.66$_{\pm0.01}$ &87.72$_{\pm0.01}$ &0.21 &-0.47 &-49.23
&88.42$_{\pm0.01}$ &88.44$_{\pm0.01}$ &1.03 &0.39 &149.75  \\
\hline
TA-SA-bi &88.38$_{\pm0.01}$ &88.41$_{\pm0.01}$ &0.79 &0.34 &91.47
&88.37$_{\pm0.01}$ &88.41$_{\pm0.01}$ &1.84 &0.33 &346.60  \\
\hline
bi-SA-TA &88.23$_{\pm0.01}$ &88.25$_{\pm0.01}$ &0.46 &0.18 &12.68
&88.63$_{\pm0.01}$ &88.65$_{\pm0.01}$ &1.61 &0.63 &291.65  \\
\hline
bi-TA-SA &88.13$_{\pm0.01}$ &88.17$_{\pm0.01}$ &0.85 &0.05 &105.64
&88.75$_{\pm0.01}$ &88.75$_{\pm0.01}$ &2.87 &0.76 &597.31  \\ \hline
\end{tabular}
} 
\end{table*}
\begin{enumerate}
    \item The $\Delta\varepsilon$ signals are denoised, detrended, and bandpass filtered (0.1–30 Hz), then segmented into overlapping 2 s windows with a 0.5 s shift and weighted by a Hamming window to reduce spectral distortion.
    \item For each window, 36 handcrafted features are extracted, including energy-based and entropy-based measures in the time and frequency domains, as well as statistical descriptors (e.g., percentile ranges) (see \cite{Fakhruzi2025} for details).
    \item Labels are segmented using the same windowing scheme to ensure alignment. For transition frames containing consecutive events, majority voting assigns the most frequent label within the frame.
    \item The 36-dimensional baseline feature set is augmented with first- and second-order temporal derivatives (+$\Delta$), yielding 108-dimensional feature vectors per frame to enhance contextual temporal representation. Their impact is analyzed in Section \ref{subsec:eff-deltas}.    
    \item The contribution of spatial context is evaluated by concatenating the features of the target SP with those of two adjacent SPs, including their corresponding +$\Delta$, yielding 324-dimensional feature vectors per frame (see Fig. \ref{fig:layout-acera-palacio}(a)). For continuous-time event classification, the resulting data are further grouped into 1.5-minute segments, which serve as input to the ML models analyzed in Section \ref{sec:comparative-rnn}.
\end{enumerate}

\subsection{Model hyperparameter selection} 
\label{subsec:optuna}
The experiments have been designed to evaluate the effectiveness of the proposed architectures and parameterizations in capturing temporal and spatial information from DAS data, as well as the contribution of attention mechanisms under the considered experimental conditions. The ablation study analyzing the interaction between attention modules and ML models is detailed in Section \ref{sec:ablation-study}. Hyperparameters have been selected based on the optimal configurations identified in the ablation study; full configurations are omitted for brevity. Bayesian optimization was performed using Optuna \cite{Akiba2019}.\\
\indent Considering the dataset size and characteristics, the hyperparameter search space for all architectural variants (with or without temporal derivatives) included: 1--3 layers, hidden size between 64 and 256, dropout rate between 0.1 and 0.3, and the Adam optimizer with a learning rate in the range $1\times10^{-5}$--$1\times10^{-4}$. L2 regularization, cross-validation, and early stopping have been applied to mitigate overfitting. A 50-trial Optuna study was conducted to ensure robustness and identify the optimal hyperparameter configurations.
\subsection{Results}
\label{subsec:experimental}
Spatial and temporal attention mechanisms within RNN architectures are evaluated using the experimental traffic dataset described in Section \ref{subsec:datadesc}. The study is structured in three stages. First, RNN variants are benchmarked to determine the optimal architecture and feature representation. Second, an ablation study analyzes the impact of attention mechanisms, examining both their placement within the network and their functional focus. Finally, the transferability of the best-performing model---trained on \textit{Palacio de Congresos} data---is assessed on traffic data from a different monitoring site, \textit{Acera del Darro} (see Table \ref{tab: urban_events}).\\
\indent All proposed architectures are optimized using Optuna (see Section \ref{subsec:optuna}) and trained with feature sets including or excluding temporal derivatives (+$\Delta$). A 5-fold cross-validation scheme ensures model stability. Performance is evaluated in terms of accuracy (Acc), F1-score (F1), number of trainable parameters (\#Param), relative improvement in accuracy over the baseline (RI-Acc), and relative parameter increase with respect to the baseline (RPI). The bi-LSTM model, selected based on the analysis in Section  \ref{sec:comparative-rnn}, serves as the baseline for all comparisons. Global results are reported in Table \ref{tab:model_comparison} and illustrated in Fig. \ref{fig:bar-all}. Table \ref{tab:model_comparison} summarizes the overall results, which will be presented again in dedicated tables in subsequent sections to enable a more detailed examination. Whereas, Fig. \ref{fig:bar-all} shows bar charts displaying the mean Acc and F1, with the y-axis restricted to the 50–90\% range for clarity and whiskers indicating the standard deviation across folds. Models are ordered from left to right by increasing trainable parameter count to provide a structured analysis of the experimental findings. The results are organized according to the three evaluation stages previously defined, as detailed below:
\begin{figure*}[!ht]
    \centering
    \includegraphics[width=17cm]{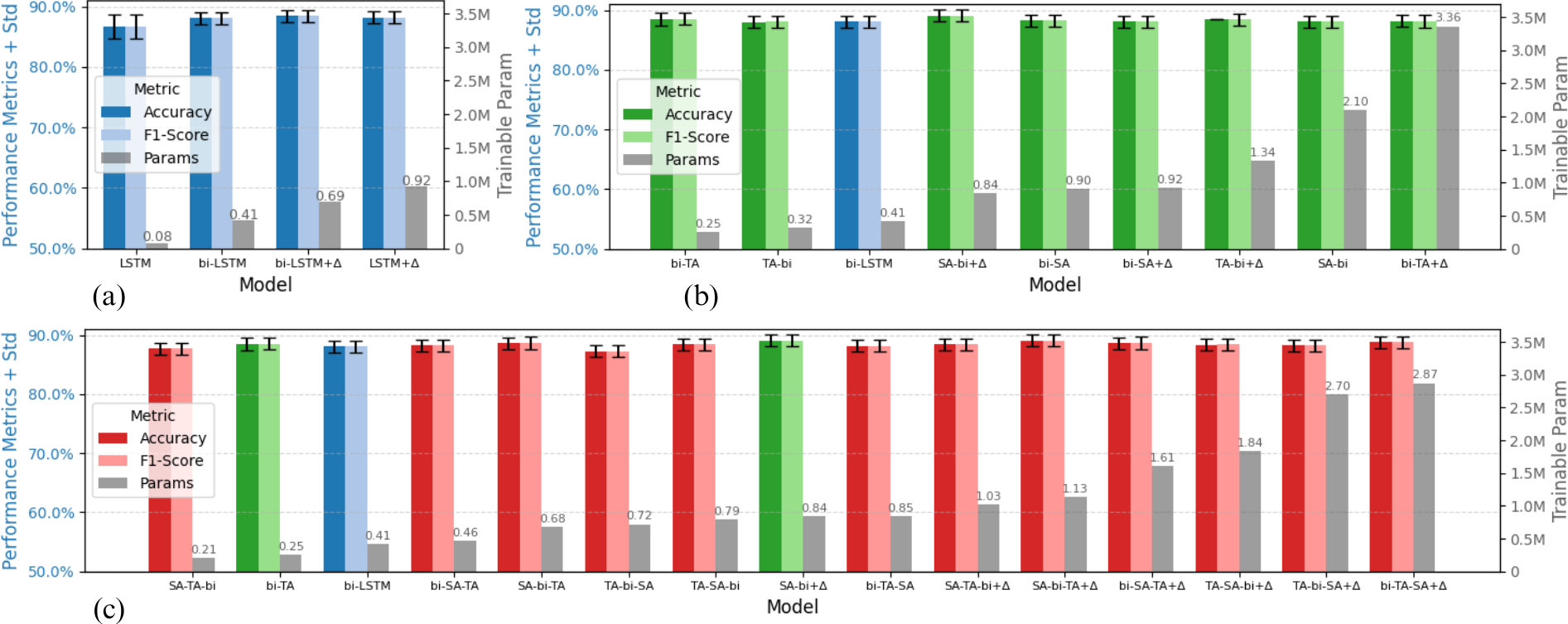}
    \caption{Model comparison in accuracy, F1-score, and standard deviation (whisker) across 5-fold cross-validation. Detail for the left-side y-axis range between 50\%--90\%. Models are sorted from left to right based on increasing number of trainable parameters (secondary y-axis) and trained with and without temporal derivatives (+$\Delta$). (a) Baseline models (blue), (b) Single attention models (green), and (c) Cascade spatio-temporal attention models (red).}
    \label{fig:bar-all}
\end{figure*}
\subsubsection{Evaluation of RNN architectures}
LSTM and bi-LSTM variants, with and without temporal derivatives, are compared to determine their effectiveness in modeling temporal dependencies. Results are reported in Table \ref{tab:model_comparison} and shown in Fig.\ref{fig:bar-all}(a).
\subsubsection{Ablation results}
\begin{figure}[!ht]
    \center\includegraphics[width=7cm]{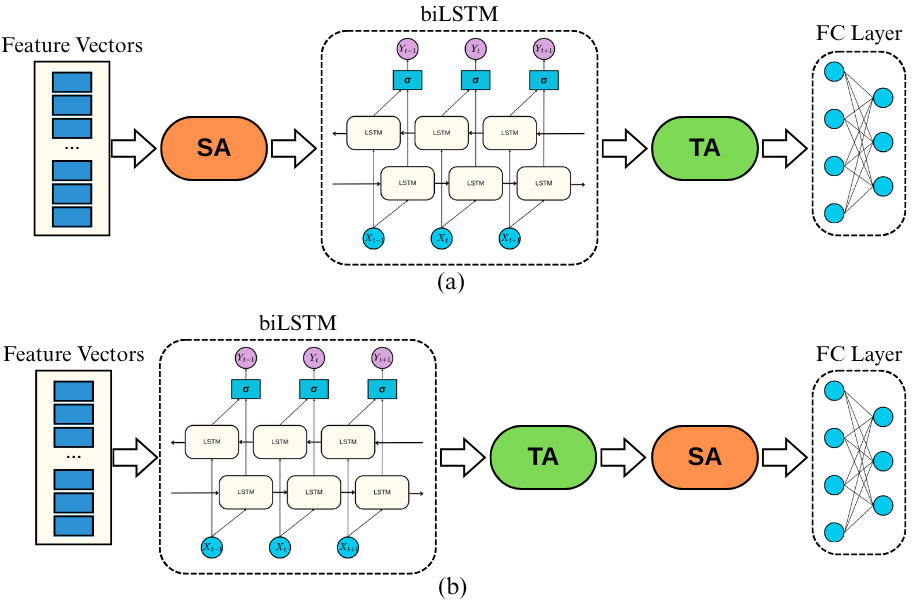}
    \caption {Two representative architectures from the ablation study: (a) SA-bi-TA, (b) bi-TA-SA.} 
    \label{fig:ablation}
\end{figure}

Starting from the bi-LSTM baseline (see Section \ref{sec:comparative-rnn}), hereafter denoted as \textit{bi}, we conduct an ablation study to evaluate the contribution of spatial attention (SA) and temporal attention (TA) modules under the proposed monitoring scenario. Two stages are considered. First, single attention configurations introduce either SA or TA before or after the baseline, yielding four architectures: SA-bi, bi-SA, TA-bi, and bi-TA. Second, sequential cascade configurations combine both SA and TA to assess their joint effect, resulting in six architectures: SA-bi-TA, TA-bi-SA, SA-TA-bi, TA-SA-bi, bi-TA-SA, and bi-SA-TA.\\
\indent Given the limited size of the labeled dataset and to avoid the additional complexity of parallel or joint designs, the analysis focuses only on sequential spatio-temporal configurations. This enables a systematic evaluation of how attention type, order, and placement influence performance, interpretability, and parameter complexity. Fig. \ref{fig:ablation} illustrates two representative architectures, while quantitative results are reported in Table \ref{tab:model_comparison} and shown in Fig. \ref{fig:bar-all}(b) and (c).
\subsubsection{Spatial transferability experiment}
\label{subsec:res-sp-transfer}
\begin{figure}[!ht]
    \center\includegraphics[width=6.5cm]{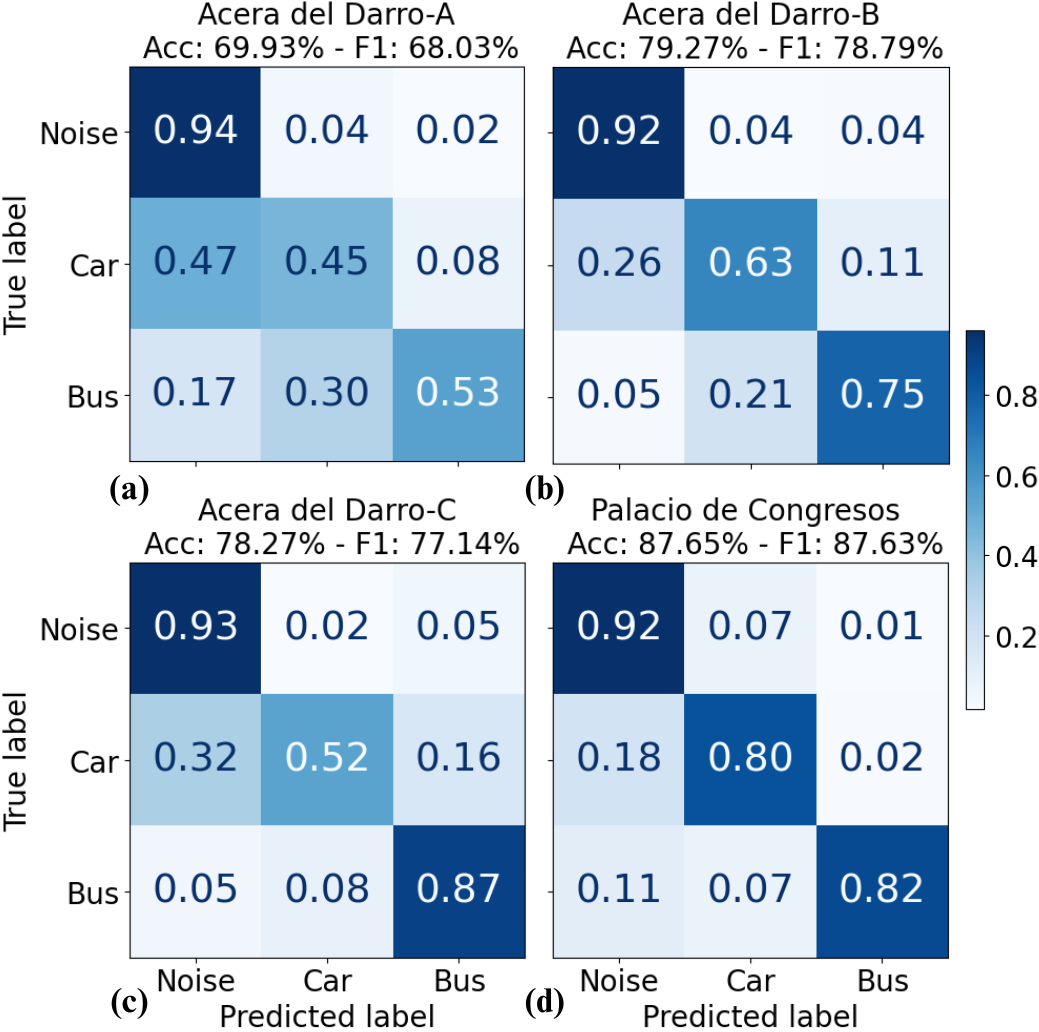}
    \caption {Confusion Matrices of the SA-bi-TA model: (a) Group A, (b) Group B, (c) Group C of \textit{Acera del Darro}, and (d) \textit{Palacio de Congresos}. } 
    \label{fig:cm-acera}
\end{figure}

To evaluate the generalization capability of the proposed architecture, we first analyze the data clustering properties at the two monitoring sites---\textit{Palacio de Congresos} and \textit{Acera del Darro} (see Fig.\ref{fig:das_map}). The automatic event recognition system described in Section \ref{sec:ablation-study}, trained on \textit{Palacio de Congresos} data, is then used to classify unseen events recorded at \textit{Acera del Darro}. The SA-bi-TA model (see Section \ref{subsec:cascade-attn} for justification) is selected for this experiment, and its performance is evaluated on the \textit{Acera del Darro} dataset. The seven SPs identified at this site are organized into three groups (A, B, and C), each comprising three SPs, to assess spatial transferability. Fig. \ref{fig:cm-acera} presents the confusion matrices of the SA-bi-TA model evaluated on Groups A, B, and C (Subfigures~(a)--(c)), alongside its evaluation on the \textit{Palacio de Congresos} test set (Subfigure~(d)).

\section{Comparative evaluation of RNN architectures}
\label{sec:comparative-rnn}
\subsection{LSTM vs. bi-LSTM} 
As shown in Table \ref{tab:lstm-bi-lstm}, the bi-LSTM consistently outperforms the LSTM in terms of accuracy, both with the original feature vectors (88.08\% vs. 86.62\%) and with temporal derivatives (88.38\% vs. 88.17\%). F1-scores exhibit the same trend across all configurations. The bi-LSTM and bi-LSTM+$\Delta$ achieve relative accuracy improvements (RI-Acc) of 1.68\% and 2.03\%, respectively, over the baseline. Exploiting both past and future context enables more effective modeling of the rapid signal transitions produced by vehicles crossing the fiber.
    \begin{table}[ht]    
    \centering
    \caption{LSTM vs. bi-LSTM comparison using feature vectors with/without temporal derivatives (+$\Delta$). Metrics: Acc, F1, \#Param, RI-Acc, and RPI. Baseline Model: LSTM (gray shading). \#Param is reported in millions (M).}
    \label{tab:lstm-bi-lstm}
    \scriptsize
    \begin{tabular}{lccp{0.8cm}p{0.8cm}r}
    \hline
    Model & Acc(\%) & F1(\%) & \#Param (M) & RI-Acc (\%) & RPI(\%) \\ 
    \hline
    \rowcolor{gray!15}
    LSTM & 86.62$_{\pm0.02}$ & 86.59$_{\pm0.02}$ & 0.08 & 0.00 & 0.00 \\ 
    \textbf{bi-LSTM} & \textbf{88.08$_{\pm0.01}$} & \textbf{88.10$_{\pm0.01}$} & \textbf{0.41} & \textbf{1.68} & \textbf{389.00} \\ 
    LSTM+$\Delta$ & 88.17$_{\pm0.01}$ & 88.18$_{\pm0.01}$ & 0.92 & 1.79 & 997.20 \\ 
    \textbf{bi-LSTM+$\Delta$} & \textbf{88.38$_{\pm0.01}$} & \textbf{88.41$_{\pm0.01}$} & \textbf{0.69} & \textbf{2.03} & \textbf{722.05} \\ 
    \hline
    \end{tabular}
    \end{table}

\subsection{Effect of temporal derivatives} 
\label{subsec:eff-deltas}
They improve LSTM performance (88.17\%, RI-Acc = 1.79\%), making it comparable to the bi-LSTM architecture (88.08\%). This indicates that short-term signal dynamics provide discriminative cues similar to those captured through bidirectional context modeling. Given the brief duration of vehicle-induced events---due to perpendicular crossings---rich short-term information enables the model to detect subtle amplitude shifts and rapid signal variations. First-order derivatives capture the rate of change, while second-order derivatives capture acceleration effects. Together, they enhance the identification of transient patterns within and between events by encoding fine-grained temporal variations. However, this performance gain comes at the expense of a substantial increase in model complexity, resulting in a relative parameter increase (RPI) of 997.20\% for LSTM+$\Delta$ and 772.05\% for bi-LSTM+$\Delta$.
\subsection{Baseline model selection} 
The inclusion of temporal derivatives triples the input dimensionality. The corresponding hyperparameter optimization yields an LSTM model with larger hidden layers, increasing the number of trainable parameters to 0.92M---an increment of 997.20\% relative to the standard LSTM. Although derivatives improve LSTM+$\Delta$ performance, the bi-LSTM achieves comparable accuracy with fewer trainable parameters (0.41M) and a narrower confidence interval. Given this favorable trade-off between performance and model complexity, the bi-LSTM is selected as the baseline architecture for incorporating spatio-temporal attention mechanisms in the subsequent ablation study.

\section{Ablation Study}
\label{sec:ablation-study}
\subsection{Single attention configurations}
\label{subsec:single-attn}
Model performance under single attention configurations is illustrated by the green bars in Fig. \ref{fig:bar-all}(b). The standard deviations of accuracy and F1-score across 5-fold cross-validation fall within comparable confidence intervals. The primary distinguishing factor is therefore the number of trainable parameters (\#Param). Consequently, model selection is guided by the trade-off between predictive performance and parameter efficiency.

    \begin{table}[ht]    
    \centering
    \caption{SA or TA module After bi-LSTM on feature vectors with/without temporal derivatives (+$\Delta$). Metrics: Acc, F1, \#Param, RI-Acc, and RPI. Baseline Model: bi-LSTM (gray shading). \#Param is reported in millions (M).}
    \label{tab:att-after}
    \scriptsize{
    \begin{tabular}{lccp{0.8cm}p{0.8cm}r}
    \hline
    Model &Acc(\%) &F1(\%) &\#Param (M) &RI-Acc (\%) &RPI(\%) \\ 
    \hline
    bi-SA   & 88.23$_{\pm0.01}$ &88.25$_{\pm0.01}$ & 0.90 &0.18 &119.50  \\ 
    \textbf{bi-TA}   & \textbf{88.47$_{\pm0.01}$} &\textbf{88.49$_{\pm0.01}$}  & \textbf{0.25} &\textbf{0.45} &\textbf{-38.85}  \\ 
    bi-SA+$\Delta$   & 88.02$_{\pm0.01}$ &88.02$_{\pm0.01}$ & 0.92 &-0.06 &123.26  \\ 
    bi-TA+$\Delta$  & 88.13$_{\pm0.01}$ &88.11$_{\pm0.01}$ & 3.36 &0.06 &715.96  \\ 
    \rowcolor{gray!15}
    bi-LSTM   & 88.08$_{\pm0.01}$ &88.10$_{\pm0.01}$ & 0.41 & 0.00 & 0.00  \\ 
    \hline
    \end{tabular}
    }
    \end{table}

\indent When SA or TA is inserted after the bi-LSTM (see Table \ref{tab:att-after}), both bi-SA (88.23\% Acc) and bi-TA (88.47\% Acc) show modest improvements over the baseline (88.08\% Acc). Their relative parameter increases (RPI) are 119.50\% and -38.85\%, respectively, underscoring the parameter efficiency of bi-TA. In contrast, when temporal derivatives are included, bi-SA+$\Delta$ (88.02\% Acc) fails to surpass the baseline. Although bi-TA+$\Delta$ reaches 88.13\% Acc (RI-Acc = 0.06\%), this marginal gain comes at the cost of a substantial RPI of 715.96\%.\\
\indent These results suggest that the temporal dynamics modeled by the bi-LSTM facilitate the subsequent attention module in emphasizing salient event patterns. In configurations without temporal derivatives, placing the attention module after the bi-LSTM compensates for the absence of explicit short-term features, yielding a more parameter-efficient architecture. This is reflected in the substantial reduction in RPI (-38.85\%) observed for the bi-TA model. Such efficiency helps mitigate overfitting while reducing computational cost.\\
\indent Conversely, inserting an SA or TA module before the bi-LSTM leads to distinct outcomes. Both TA-bi (87.97\% Acc) and SA-bi (88.07\% Acc) perform within the same range as the baseline bi-LSTM (88.08\% Acc), with the main difference reflected in their relative parameter increase (RPI) (see Table \ref{tab:att-before}). When temporal derivatives are included, TA-bi+$\Delta$ (88.41\%) and SA-bi+$\Delta$ (89.05\%) show modest accuracy improvements. Notably, SA-bi+$\Delta$ achieves the highest RI-Acc (1.10\%), albeit at the cost of a substantial RPI of 103.50\%, revealing a clear performance–complexity trade-off. This configuration may therefore be preferable in scenarios where predictive accuracy is prioritized over computational efficiency.\\
\indent The results in Section \ref{subsec:single-attn} show that, beyond the intrinsic temporal modeling capability of the bi-LSTM, comparable accuracy can be achieved through three alternative strategies with different complexity levels: (i) incorporating temporal derivatives into the input features without attention; (ii) appending a TA module after the bi-LSTM; and (iii) combining temporal derivatives with an attention module placed before the bi-LSTM. From a complexity standpoint, TA notably reduces the number of trainable parameters while maintaining similar accuracy. In contrast, SA alone does not yield measurable improvements. However, when combined with temporal derivatives (SA-bi+$\Delta$), modest yet consistent accuracy gains are obtained at the expense of increased model complexity. 

    \begin{table}[ht]    
    \centering
    \caption{SA or TA module Before bi-LSTM on feature vectors with/without temporal derivatives (+$\Delta$). Metrics: Acc, F1, \#Param, RI-Acc, and RPI. Baseline Model: bi-LSTM (gray shading). \#Param is reported in millions (M).}
    \label{tab:att-before}
    \scriptsize{
    \begin{tabular}{lccp{0.8cm}p{0.8cm}r}
    \hline
    Model & Acc(\%) & F1(\%) & \#Param (M) & RI-Acc (\%) & RPI(\%) \\  
    \hline
    SA-bi   & 88.07$_{\pm0.01}$ &88.09$_{\pm0.01}$ & 2.10 &-0.01 &408.64  \\ 
    TA-bi   & 87.97$_{\pm0.01}$ &88.06$_{\pm0.01}$ & 0.32 &-0.12 &-23.21  \\ 
    \textbf{SA-bi+$\Delta$}   & \textbf{89.05$_{\pm0.01}$} &\textbf{89.05$_{\pm0.01}$}  & \textbf{0.84} &\textbf{1.10} &\textbf{103.50}  \\ 
    TA-bi+$\Delta$  & 88.41$_{\pm0.00}$ &88.45$_{\pm0.01}$ & 1.34 &0.37 &224.72   \\
    \rowcolor{gray!15}
     bi-LSTM   & 88.08$_{\pm0.01}$ &88.10$_{\pm0.01}$  & 0.41 & 0.00 & 0.00  \\ 
    \textbf{bi-TA}   & \textbf{88.47$_{\pm0.01}$} &\textbf{88.49$_{\pm0.01}$}  & \textbf{0.25} &\textbf{0.45} &\textbf{-38.85}  \\ 
    \hline
    \end{tabular}
    }
    \end{table}

\subsection{Cascade SA and TA configurations}
\label{subsec:cascade-attn}
Tables \ref{tab:Two-after} and \ref{tab:Two-before} compare cascade attention configurations with the two best-performing setups identified in Section \ref{subsec:single-attn}. Stacking two consecutive attention modules before or after the bi-LSTM does not provide performance gains. In contrast, single attention configurations achieve comparable or slightly higher RI-Acc while requiring fewer parameters (see Fig.~\ref{fig:bar-all}(c)). Furthermore, incorporating temporal derivatives yields only marginal accuracy improvements relative to SA-bi+$\Delta$, but at the expense of a substantial increase in model complexity (1--2.8M \#Param). Overall, cascade attention mechanisms do not outperform single attention setups, indicating diminishing returns from increased architectural depth.
    \begin{table}[!ht]    
    \centering
    \caption{SA and TA After bi-LSTM on feature vectors with/without temporal derivatives (+$\Delta$). Metrics: Acc, F1, \#Param, RI-Acc, and RPI. Baseline Model: bi-LSTM (gray shading). \#Param is reported in millions (M).}
    \label{tab:Two-after}
    \scriptsize{
    \begin{tabular}{lccp{0.8cm}p{0.8cm}r}
    \hline
    Model & Acc(\%) & F1(\%) & \#Param (M) & RI-Acc (\%) & RPI(\%) \\ 
    \hline
    bi-SA-TA   &88.23$_{\pm0.01}$ &88.25$_{\pm0.01}$  & 0.46 &0.18 &12.68  \\ 
    bi-TA-SA   &88.13$_{\pm0.01}$ &88.17$_{\pm0.01}$  & 0.85 &0.05 &105.64  \\ 
    bi-SA-TA+$\Delta$  &88.63$_{\pm0.01}$ &88.65$_{\pm0.01}$ & 1.61 &0.63 &291.65  \\ 
    \textbf{bi-TA-SA+$\Delta$}  &\textbf{88.75$_{\pm0.01}$} &\textbf{88.75$_{\pm0.01}$} & \textbf{2.87} &\textbf{0.76} &\textbf{597.31}  \\
    \rowcolor{gray!15}
    bi-LSTM   & 88.08$_{\pm0.01}$ &88.10$_{\pm0.01}$  & 0.41 & 0.00 & 0.00  \\
    \textbf{bi-TA}   & \textbf{88.47$_{\pm0.01}$} &\textbf{88.49$_{\pm0.01}$}  & \textbf{0.25} &\textbf{0.45} &\textbf{-38.85}  \\ 
    \textbf{SA-bi+$\Delta$}   & \textbf{89.05$_{\pm0.01}$} &\textbf{89.05$_{\pm0.01}$}  & \textbf{0.84} &\textbf{1.10} &\textbf{103.50}  \\ 
    \hline
    \end{tabular}
    }
    \end{table}
    \begin{table}[!ht]    
    \centering
    \caption{SA and TA Before bi-LSTM on feature vectors with/without temporal derivatives (+$\Delta$). Metrics: Acc, F1, \#Param, RI-Acc, and RPI. Baseline Model: bi-LSTM (gray shading). \#Param is reported in millions (M).}
    \label{tab:Two-before}
    \scriptsize{
    \begin{tabular}{lccp{0.8cm}p{0.8cm}r}
    \hline
    Model & Acc(\%) & F1(\%) & \#Param (M) & RI-Acc (\%) & RPI(\%) \\  
    \hline
    TA-SA-bi   &88.38$_{\pm0.01}$ &88.41$_{\pm0.01}$ & 0.79 &0.34 &91.47  \\ 
    SA-TA-bi   &87.66$_{\pm0.01}$ &87.72$_{\pm0.01}$ & 0.21 &-0.47 &49.23  \\ 
    TA-SA-bi+$\Delta$  &88.37$_{\pm0.01}$ &88.41$_{\pm0.01}$ & 1.84 &0.33 &346.60 \\
    \textbf{SA-TA-bi+$\Delta$}  &\textbf{88.42$_{\pm0.01}$} &\textbf{88.44$_{\pm0.01}$} & \textbf{1.03} &\textbf{0.39} &\textbf{149.75}  \\ 
    \rowcolor{gray!15}
    bi-LSTM   & 88.08$_{\pm0.01}$ &88.10$_{\pm0.01}$  & 0.41 & 0.00 & 0.00  \\
    \textbf{bi-TA}   & \textbf{88.47$_{\pm0.01}$} &\textbf{88.49$_{\pm0.01}$}  & \textbf{0.25} &\textbf{0.45} &\textbf{-38.85}  \\ 
    \textbf{SA-bi+$\Delta$}   & \textbf{89.05$_{\pm0.01}$} &\textbf{89.05$_{\pm0.01}$}  & \textbf{0.84} &\textbf{1.10} &\textbf{103.50}  \\ 
    \hline
    \end{tabular}
    }
    \end{table}
\indent Table \ref{tab:between-Attn} presents the results obtained by inserting the bi-LSTM between attention modules. Accuracy remains within a comparable range, indicating that positioning SA before the bi-LSTM and TA afterward constitutes a reasonable design choice. As expected, incorporating temporal derivatives increases \#Param. The highest RI-Acc (1.10\%) is achieved by SA-bi-TA+$\Delta$, with an RPI of 174.15\%, slightly exceeding that of SA-bi+$\Delta$ (RPI = 103.50\%, RI-Acc = 1.10\%). This indicates that temporal derivatives in SA-bi+$\Delta$ capture temporal dynamics more efficiently, reinforcing the observation of diminishing returns when combining multiple temporal modeling mechanisms.\\
\indent Key insights from the cascade attention analysis indicate that the highest RI-Acc (1.10\%), achieved by SA-bi-TA+$\Delta$, can be obtained more efficiently using a single SA module combined with temporal derivatives (SA-bi+$\Delta$). Moreover, Section \ref{subsec:expl-two-attn} shows that SA-bi-TA without temporal derivatives achieves a comparable RI-Acc (0.62\%) with a substantially lower RPI (65.48\%), while offering improved interpretability. In particular, its attention weights highlight distinct spatial and temporal patterns associated with traffic events, providing insight into the model’s decision process and suggesting a favorable trade-off among performance, complexity, and interpretability. 

    \begin{table}[!ht]    
    \centering
    \caption{bi-LSTM between SA and TA on feature vectors with/without temporal derivatives (+$\Delta$). Metrics: Acc, F1, \#Param, RI-Acc, and RPI. Baseline Model: bi-LSTM (gray shading). \#Param is reported in millions (M).}
    \label{tab:between-Attn}
    \scriptsize{
    \begin{tabular}{lccp{0.8cm}p{0.8cm}r}
    \hline
    Model & Acc(\%) & F1(\%) & \#Param (M) & RI-Acc (\%) & RPI(\%) \\ 
    \hline
    TA-bi-SA   &87.24$_{\pm0.01}$ &87.28$_{\pm0.01}$  & 0.72 &-0.95 &88.20 \\ 
    \textbf{SA-bi-TA}   &\textbf{88.62$_{\pm0.01}$} &\textbf{88.65$_{\pm0.01}$}  & \textbf{0.68} &\textbf{0.62} &\textbf{65.48} \\ 
    TA-bi-SA+$\Delta$  &88.20$_{\pm0.01}$ &88.22$_{\pm0.01}$ & 2.70 &0.14 &555.32 \\ 
    \textbf{SA-bi-TA+$\Delta$}  &\textbf{89.05$_{\pm0.01}$} &\textbf{89.03$_{\pm0.01}$} & \textbf{1.13} &\textbf{1.10} &\textbf{174.15}  \\
    \rowcolor{gray!15}
    bi-LSTM   & 88.08$_{\pm0.01}$ &88.10$_{\pm0.01}$  & 0.41 & 0.00 & 0.00  \\
    \textbf{bi-TA}   & \textbf{88.47$_{\pm0.01}$} &\textbf{88.49$_{\pm0.01}$}  & \textbf{0.25} &\textbf{0.45} &\textbf{-38.85}  \\ 
    \textbf{SA-bi+$\Delta$}   & \textbf{89.05$_{\pm0.01}$} &\textbf{89.05$_{\pm0.01}$}  & \textbf{0.84} &\textbf{1.10} &\textbf{103.50}  \\
    \hline
    \end{tabular}
    }
    \end{table}

\begin{figure*}
    \center\includegraphics[width=16cm]{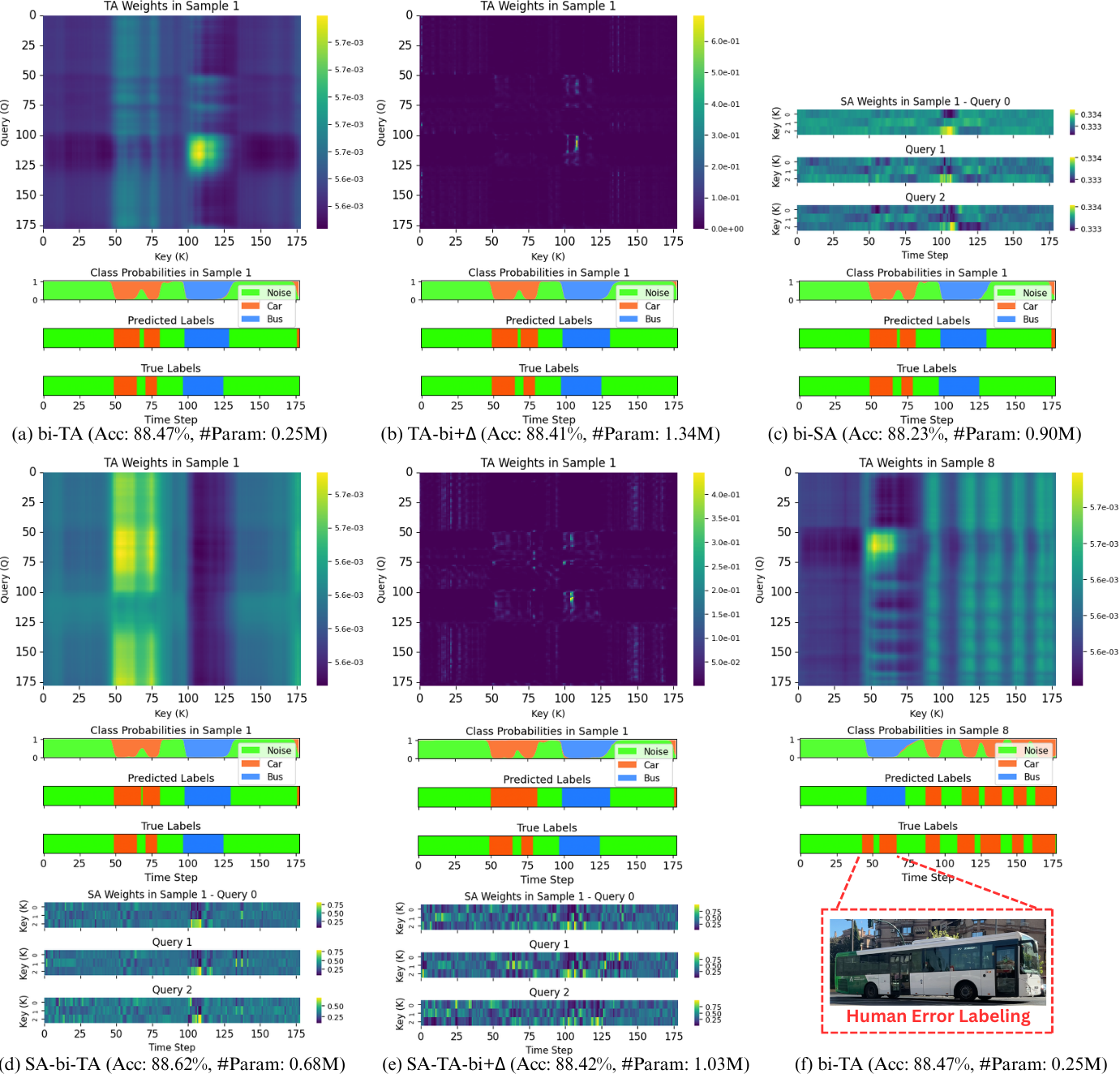}
    \caption{Spatial and temporal attention weight heatmaps from several models with their model's class probabilities output, predicted labels, and ground truth labels (True Labels) over the sequence: (a) bi-TA, (b) TA-bi+$\Delta$, (c) bi-SA,  (d) SA-bi-TA, (e) SA-TA-bi+$\Delta$, (f) bi-TA with a human error labeling identification.}
    \label{fig:attn-weights}
\end{figure*}

\section{Attention Weights Visualization}
\label{sec:att-weights}
\indent The resulting architectures are further analyzed through attention weights visualization to gain insight into the models’ decision-making processes and provide qualitative evidence of the learned spatial and temporal dependencies. Attention weights are computed by measuring the similarity between the query ($Q$) and key ($K$) vectors (Equation 1). In TA, $Q$ determines which temporal windows within a sequence should be emphasized to extract relevant information, while $K$ encodes the content of each window (i.e., the corresponding vehicle signature). Similarly, in SA, $Q$ identifies the SPs that should be prioritized within each window, whereas $K$ represents the information contained in each SP. Fig. \ref{fig:attn-weights} presents the attention weight heatmaps from several representative models evaluated on the same time sequence, except for Subfigure~(f), which corresponds to a different sequence. The SA and TA heatmaps are accompanied by the corresponding class probabilities, predicted labels, and ground-truth labels (true labels) across the sequence.

\subsection{Attention maps for single attention configurations}
\label{subsec:expl-one-attn}
Fig. \ref{fig:attn-weights}(a) presents the temporal attention heatmap produced by the bi–TA model. The map reveals a strong alignment between the $Q$ and $K$ vectors within the window range 95--126, indicating that this segment constitutes a salient portion of the sequence. According to the ground-truth labels, these windows correspond to the \textit{Bus} class. Additionally, consistent moderate attention is observed from all $Q$ positions toward the $K$ vectors in windows 46--64 and 72--80, which are associated with the \textit{Car} class. This suggests that \textit{Car} events generate a distinct yet more distributed attention pattern compared to the more concentrated focus observed for the \textit{Bus}. In contrast, the \textit{Noise} class receives comparatively lower attention. The continuity of the attention distribution along the $K$-axis within these windows reflects the temporal progression of the event sequences. Such behavior indicates that the temporal dependencies initially captured by the bi-LSTM are effectively reinforced by the subsequent TA module, enabling clearer emphasis on the identified traffic events. Importantly, these attention patterns align with the temporal structure expected for real traffic events.\\
\indent Placing TA module before the bi-LSTM (TA-bi) results in an attention distribution that differs from that observed in the bi–TA configuration. Fig. \ref{fig:attn-weights}(b) presents the TA–bi+$\Delta$ heatmap, where attention weights are more concentrated on a limited set of windows. Although the highlighted regions correspond to the identified classes, the attention appears less continuous across the event sequences and does not clearly reflect their temporal progression, in contrast to the bi–TA model. This behavior likely stems from positioning the TA module at the input stage, where it operates on raw feature representations without access to the higher-level temporal context later captured by the bi-LSTM. Consequently, early-stage attention appears to serve a complementary role by emphasizing a small number of salient windows, thereby facilitating the bi-LSTM’s subsequent modeling of temporal dependencies. It suggests that the attention placement critically influences the type of temporal information captured by the architecture.\\
\indent Unlike temporal attention heatmaps, SA visualizations represent the interactions between the $Q$ and $K$ vectors across three spatial points (SPs 0--2), illustrating how specific locations are weighted for different classes throughout the sequence. Fig. \ref{fig:attn-weights}(c) presents the heatmap of the bi–SA model, where attention is strongly allocated to the $K$ vectors of SP 2 within windows 98--110. According to the ground-truth labels, these windows correspond to the \textit{Bus} class, indicating that the bi-SA configuration selectively prioritizes SP 2 for bus identification. This observation is consistent with the physical layout of the monitored roadway, as SP 2 is positioned along the lane predominantly used by buses. However, the heatmap does not reveal clearly distinguishable attention structures for the \textit{Car} and \textit{Noise} classes.
\subsection{Attention maps for cascade attention configurations} 
\label{subsec:expl-two-attn}
The heatmaps of the sequential attention configurations, SA-bi-TA (Fig. \ref{fig:attn-weights}(d)) and SA-TA-bi+$\Delta$ (Fig. \ref{fig:attn-weights}(e)), exhibit distinct attention distributions related to the placement of the TA module, consistent with the behavior observed in the single attention configurations (see Section \ref{subsec:expl-one-attn}). When positioned after the bi-LSTM, TA reinforces the temporal dependencies already captured by the recurrent layer. In contrast, when applied before the bi-LSTM, TA operates on raw feature representations with limited contextual information, leading to a more selective emphasis on a small set of key windows.\\
\indent Notably, the SA-bi-TA configuration exhibits a complementary interplay between spatial and temporal attention mechanisms. The SA module assigns higher weights to SP 2 for the \textit{Bus} class, consistent with its physical location along the lane predominantly used by buses. The subsequent TA module emphasizes the temporal patterns associated with the \textit{Car} and, to a lesser extent, the \textit{Noise} classes, resulting in an attention distribution that differs from that observed in the bi-TA model.\\
\indent Taken together, these observations suggest a hierarchical attention process in which SA first identifies the most informative locations, while TA then models the contextual evolution of the events. This complementary behavior enhances the interpretability of the SA-bi-TA architecture by providing a coherent visual account of the spatial and temporal factors underlying its predictions.

\subsection{Attention-guided analysis of a labeling inconsistency}
While the previous examples illustrate successful event recognition, attention weights visualizations can also reveal potential labeling inconsistencies, as shown in Fig. \ref{fig:attn-weights}(f), which corresponds to a different time sequence analyzed using the bi-TA configuration. The heatmap indicates a focused alignment between the $Q$ and $K$ vectors within windows 48--52, resembling the distribution observed in Fig. \ref{fig:attn-weights}(a) and typically associated with the \textit{Bus} class. However, these windows were manually annotated as \textit{Car}, whereas the model predicts \textit{Bus}.\\
\indent To investigate this discrepancy, the corresponding video recordings from the data acquisition stage were reviewed, confirming that the identified windows indeed correspond to a bus event. This observation highlights two key aspects: (i) the model appears capable of learning discriminative attention structures associated with specific classes; and (ii) such structures can assist in identifying potential annotation inconsistencies within the dataset. A similar behavior is observed in the SA-bi-TA configuration, further supporting the consistency of the learned attention patterns. Overall, this example illustrates the potential of attention-based analysis as a complementary tool for assessing dataset quality.

\section{Spatial Transferability Analysis}
\label{sec:spat-trans}
The spatial transferability experiment described in Section \ref{subsec:res-sp-transfer} evaluates the ability of the models to recognize events at sensing locations different from those used during training. This task is inherently challenging, as each SP exhibits distinct transmission characteristics influenced by factors such as soil properties, ground–fiber coupling conditions, and noise contributions from adjacent traffic lanes. Moreover, the relative trajectory of vehicles with respect to the fiber deployment introduces additional variability in the recorded signals. For instance, vehicles crossing the fiber perpendicularly generate shorter temporal footprints than those traveling parallel to it. These factors collectively complicate the generalization of models across sensing locations.
\subsection{Cluster structure comparison}
\label{subsub:cluster}
\begin{figure}[!ht]
    \center\includegraphics[width=8.5cm]{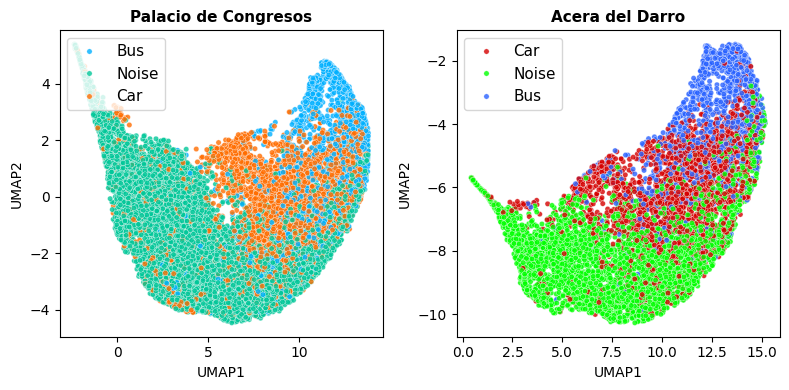}
    \caption{2-D UMAP projections of window-based feature representations (36 features) extracted from one hour of traffic data at (left) \textit{Palacio de Congresos} and (right) \textit{Acera del Darro}.} 
    \label{fig:cluster-structure}
\end{figure}

To examine the similarity between the data distributions of the two collected datasets (\textit{Palacio de Congresos} and \textit{Acera del Darro}), we employed the widely used dimensionality reduction technique Uniform Manifold Approximation and Projection (UMAP) \cite{McInnes2018}. 2-D projections were generated from the 36-feature vectors extracted from one hour of recordings at each site. As shown in Fig. \ref{fig:cluster-structure}, the embeddings were computed independently for each dataset; therefore, the comparison is qualitative and based on the observed cluster structures. While not intended for direct geometric comparison, these projections provide insight into the structural organization of each dataset, enabling the characterization of the underlying distributions and the assessment of clustering tendencies and class separability.\\
\indent The projections reveal broadly comparable distributions, suggesting a meaningful degree of structural similarity across the two locations. In both UMAP embeddings, the \textit{Car} class exhibits noticeable overlap with the \textit{Bus} and \textit{Noise} classes, although partial cluster separation remains observable. These observations indicate that, despite moderate inter-class overlap, the datasets retain sufficient structural coherence to support cross-site model transfer. Nevertheless, some degree of site-specific adaptation may be required to further improve performance. Overall, the results support the feasibility of spatial transfer while highlighting the importance of accounting for site-dependent variability.

\subsection{Classification results}
\label{subsub:general}
To evaluate model transferability, the SA-bi-TA configuration trained on the \textit{Palacio de Congresos} dataset has been applied to the unseen \textit{Acera del Darro} dataset. The SPs identified at \textit{Acera del Darro} were organized into 3 groups A (\#1--\#3), B (\#3--\#5) and C (\#5--\#7), with different spatial properties, monitoring the four traffic lanes with two traffic directions depicted in Fig. \ref{fig:layout-acera-palacio}(b).\\
\indent Fig. 6 presents the confusion matrices of the SA-bi-TA model for each group of the Acera del Darro dataset (Fig. \ref{fig:cm-acera}(a)--(c)) and for the \textit{Palacio de Congresos} dataset (Fig. \ref{fig:cm-acera}(d)). Applying the trained model to a different location results in a decrease in global average accuracy of up to 17.72\% (\textit{Palacio de Congresos} vs. \textit{Acera del Darro} Group A). Among the three \textit{Acera del Darro} groups, Group B achieves the highest performance, with an accuracy of 79.27\%, compared to Group A (69.93\%) and Group C (78.27\%). This improved performance is likely related to the central positioning of the corresponding SPs along lanes 2 and 3 (see Fig. \ref{fig:layout-acera-palacio}(b)), where coupling conditions are more favorable and proximity to vehicles crossing the fiber is higher despite the surrounding noise. Regarding per-class accuracy, all the scenarios have similar behavior. \textit{Bus} results slightly vary across groups, while \textit{Noise} is consistently well-identified and \textit{Car} remains the most misclassified category.

\section{Insights from Experimental Results}
\label{sec:insights}
The analyses presented in the previous sections indicate that spatial and temporal information derived from DAS measurements provide complementary contributions to traffic event recognition. RNN-based models effectively capture the sequential nature of events. While standard LSTM architectures achieve satisfactory performance, the bi-LSTM offers the best balance between accuracy, number of trainable parameters, and stability across 5-fold cross-validation, and was therefore selected as the baseline architecture for the ablation study of spatial and temporal attention mechanisms.\\
\indent Although the performance of attention-enhanced models is generally comparable to that of the baseline bi-LSTM, the ablation study provides important architectural insights, particularly regarding parameter efficiency. TA captures temporal dependencies more efficiently than temporal derivatives (+$\Delta$), achieving similar performance with a lower number of trainable parameters. This suggests a more effective utilization of model capacity to represent the underlying dynamics of the data. Parameter-efficient architectures help mitigate overfitting, which is especially relevant in real-world sensing scenarios characterized by variability and noise. Furthermore, reducing model complexity lowers computational demands during training and inference, facilitating scalability across monitoring sites and supporting practical deployment in continuous traffic monitoring systems. SA, in turn, aligns with roadway geometry, providing information that can be related to lane usage patterns. Overall, improved predictive performance does not necessarily require increased architectural complexity, but rather an effective representation of relevant spatial and temporal patterns.\\
\indent Beyond reducing model complexity, attention mechanisms provide a degree of interpretability through the visualization of attention weights, whose behavior can be directly related to the physical phenomena underlying the recorded signals. Both single (SA or TA) and cascade attention (SA-TA) configurations reveal consistent principles. Performance improvements over the baseline bi-LSTM are achieved when sufficient temporal context is available, either through the hidden representations of the bi-LSTM or through temporal derivatives. The placement of attention modules within the architecture influences their functional role. When positioned after the bi-LSTM, attention emphasizes higher-level temporal context extracted by the recurrent layers. When positioned before the bi-LSTM, temporal derivatives become necessary to provide explicit short-term context, enabling effective modeling at the expense of increased parameter count. Among the evaluated configurations, the SA-bi-TA architecture offers enhanced interpretability, as its attention heatmaps clearly highlight both spatial and temporal characteristics of traffic events.\\
\indent The spatial transferability analysis done provides key insights for the usage of DAS technology in real scenarios. The results demonstrate that, for the three canonical traffic event classes considered in this study, a model trained using data from a single sensing location can successfully recognize events occurring at different spatial locations, albeit with a moderate reduction in accuracy. This finding is particularly significant from an operational perspective, as annotating data independently for every sensing point in a distributed fiber deployment is often impractical and resource-intensive. The ability to transfer learned representations across locations therefore represents a key step toward scalable DAS-based traffic monitoring, reducing the reliance on extensive site-specific labeling while maintaining reliable event recognition performance. The observed ability of the model to generalize across sensing locations resonates with the emerging paradigm of foundation models, where learning transferable representations reduces the need for extensive task- or location-specific supervision.

\section{Conclusions}
\label{sec:conc}
This work has explored the use of spatio-temporal attention mechanisms within recurrent neural networks for distributed acoustic sensing (DAS)-based traffic monitoring under real urban conditions in Granada, Spain. From a conceptual perspective, the effectiveness of attention mechanisms in this setting can be attributed to their ability to selectively emphasize informative regions of the DAS signal while attenuating variability arising from noise, heterogeneous coupling conditions, and unrelated background activity. Traffic events manifest as localized spatio-temporal patterns whose relevance varies across sensing points and time. By dynamically weighting spatial locations and temporal segments, attention mechanisms naturally align with the structured yet heterogeneous nature of DAS measurements, enabling the model to focus on physically meaningful signal components that drive event recognition.\\
\indent Beyond demonstrating the feasibility of attention-enhanced models for automatic traffic event recognition, the study highlights the importance of jointly considering sensing physics, model design, and deployment constraints when developing learning-based monitoring systems. The results suggest that combining temporal modeling with attention mechanisms enables robust event recognition while offering meaningful interpretability through attention visualizations. In addition, the observed ability to transfer models across sensing locations points toward the possibility of learning representations that capture underlying signal structures shared across heterogeneous environments, an essential property for scaling DAS monitoring systems without requiring exhaustive site-specific annotation.\\
\indent At the same time, the findings underscore several open challenges. Variability in coupling conditions, traffic patterns, and environmental noise continues to limit cross-site performance, indicating the need for adaptive learning strategies capable of accommodating changing sensing conditions. Furthermore, the large volume of DAS data and the cost of manual labeling remain significant bottlenecks for operational deployment. These limitations point at research directions for finding more more scalable, reliable, and interpretable DAS-based traffic monitoring frameworks, supporting the broader vision of intelligent transportation systems that can continuously learn from real-world sensing environments.

\bibliographystyle{IEEEtran}
\bibliography{references}

\end{document}